\pdfoutput=1
\documentclass[letterpaper, 10 pt, conference]{ieeeconf}  

\IEEEoverridecommandlockouts                              

\usepackage{caption}
\usepackage{subfigure}
\usepackage{cite}
\usepackage{amsmath,amssymb,amsfonts}
\usepackage{graphicx}
\usepackage{textcomp}
\usepackage{xcolor}
\usepackage{booktabs}
\usepackage{multirow}
\usepackage{amsfonts,amssymb}
\usepackage[algo2e]{algorithm2e}
\usepackage{algorithm}  
\usepackage{algpseudocode}  
\usepackage{booktabs} 
\usepackage{amsmath}  

\def\BibTeX{{\rm B\kern-.05em{\sc i\kern-.025em b}\kern-.08em
    T\kern-.1667em\lower.7ex\hbox{E}\kern-.125emX}}

\title{\LARGE \bf
Joint State and Input Estimation of Agent Based on Recursive Kalman Filter Given Prior Knowledge
}

\author{Zida Wu$^{1}$,  Zhaoliang Zheng $^{1}$, and Ankur Mehta$^{1}$ 
\thanks{ $^{1}$Department of Electrical and Computer Engineering, University of California Los Angeles, Los Angeles, CA, USA.
        {\tt\small \{zdwu,zhz03,mehtank\} @ucla.edu}        }%
}

\begin{document}

\maketitle
\thispagestyle{empty}
\pagestyle{empty}

\begin{abstract}

Modern autonomous systems are purposed for many challenging scenarios, where agents will face unexpected events and complicated tasks. The presence of disturbance noise with control command and unknown inputs can negatively impact robot performance. Previous research of joint input and state estimation separately studied the continuous and discrete cases without any prior information. This paper combines the continuous and discrete input cases into a unified theory based on the Expectation-Maximum (EM) algorithm. By introducing prior knowledge of events as the constraint, inequality optimization problems are formulated to determine a gain matrix or dynamic weights to realize an optimal input estimation with lower variance and more accurate decision-making. Finally, statistical results from experiments show that our algorithm owns 81\% improvement of the variance than KF and 47\% improvement than RKF in continuous space; a remarkable improvement of right decision-making probability of our input estimator in discrete space, identification ability is also analyzed by experiments.

\end{abstract}

\section{INTRODUCTION}
\label{section1}

Unknown input estimation has received a great deal of attention over past decades with implementation in numerous applications, such as geophysical detection \cite{taher2021earthquake}, environment monitoring \cite{2021Decentralized, kitanidis1987unbiased}, sensor fault detection and diagnosis \cite{lu2015double}, intention inference for self-driving \cite{2016Simultaneous}, and maneuvering tracking \cite{1979A, 2009Modified, 2015IMM}, among others. 
\label{section1}
\begin{figure}[htb]        
 \center{\includegraphics[width=7cm]  {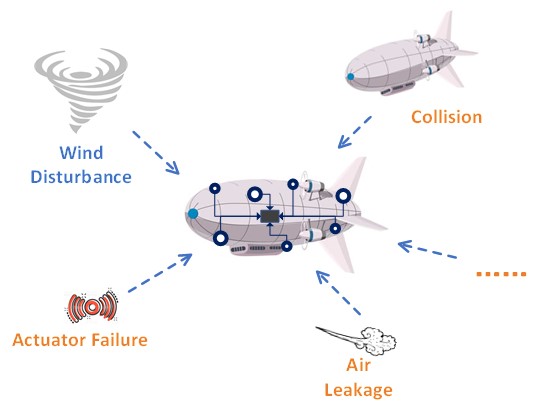}}        
\caption{\label{1} An inspection blimp may encounter many unexpected events, regardless of whether it is natural disaster event detection or an unexpected disturbance on agents, these events can all be regarded as inputs acting on dynamic systems.}  
\setlength{\belowcaptionskip}{-1cm} 
\vspace{-0.4cm} 
 \end{figure}

Given the dynamics of inputs, a popular approach called the augmented Kalman filter\cite{2012An} augments input parameters into a state vector. Due to exploding computational cost as the state dimension increases, a two-stage Kalman filter \cite{2000Robust} was proposed to reduce complexity by using two sub-filters to decouple state and input. In practice, however, the input dynamics are often unknown, and should be treated accordingly as an unexpected disturbance on the system. Kitanidis \cite{kitanidis1987unbiased} and Darouach \cite{darouach1997unbiased} first developed optimal state filters when the prior information of the input was unavailable. Hsieh\cite{2000Robust} unified the two-stage filter and the optimal filter in \cite{kitanidis1987unbiased}, where inputs were also estimated. Afterward, Gillijns \cite{gillijns2007unbiased} proposed a joint unbiased and minimum-variance (MVU) estimation of input and state. Furthermore, he theoretically proved the optimality of the recursive Kalman filter (RKF) when satisfying rank conditions. In later research, \cite{darouach2003extension, gillijns2007unbiasedb, 2009Extension} extended the problem when the input has direct feedthrough, which means the unknown input affects both state and output. Later on, many works proposed relaxing the rank conditions of the system\cite{2010On, 2016A, Hsieh2017Unbiased, kong2019internal}. Most recently, some researchers focused on a more general and optimal estimation when multi-step delayed and simultaneously relaxed the rank requirement of feedthrough \cite{2017Multi, 0Time, 9289795}.  

So far, the input estimation problem in continuous space has been well studied by various kinds of solutions. However, this previous research deals with input without any prior knowledge. On this condition, input-state estimation is achieved but at the cost of large variance. Input-state estimation under a prior constraint is studied in \cite{kong2021kalman}, but only the norm constraint is discussed.

In this paper, we extend previous input estimation into a general input-state joint estimation problem with prior knowledge, where the input information is partially known before taking action. Such prior knowledge, as an extra constraint, will limit the range of input and lead to a lower variance. Furthermore, we combine both continuous and discrete input cases into a unified theory based on Expectation-Maximum (EM) algorithm. For the continuous space, we mainly improve the input estimation step from the original RKF \cite{gillijns2007unbiased} and its variants. An inequality constraint problem is formulated to find the optimal gain matrix and solved by the augmented Lagrangian method. As for the discrete input case, we refer to the multi-models (MM) idea which is a popular solution to maneuver tracking or driving intentions inference when a potential motion mode is available \cite{2016Simultaneous}\cite{pitre2005comparative}.

Our approach provides a general solution to estimate an unexpected disturbance or impulse force acting on a dynamic system. By introducing prior knowledge of the target, our algorithm can handle various kinds of input-based estimation tasks well. 


\section{Problem Formulation}
\label{section2}
\subsection{System description}\label{AA}
\begin{equation}
{x_{k + 1}} = {A_k}{x_k} + {G_k}{d_k} + {w_k} \label{eq1}
\end{equation}

\begin{equation}
{y_{k+1}} = {C_{k + 1}}{x_{k + 1}} + {v_{k+1}} \label{eq2}
\end{equation}

where $x \in {\mathbb{R}^n}$ is the state, $d \in {\mathbb{R}^m}$ is the input, $y \in {\mathbb{R}^p}$ is the measurement of the system. The process noise is $w$ while the measurement noise is $v$, which  are supposed as uncorrelated, zero-mean, white Gaussian noises. 


This paper considers the problem of the input and state joint estimation of a linear time-varying system in the presence of prior knowledge of input. In the traditional Kalman filter, the value of inputs $d_k$ as commanded actions is specifically known. However, in practice, inputs are often actuated by unknown drivers which can be regarded as an uncontrolled action. This uncontrolled input can be accounted for the process noise but would lead to a large variance of estimation. The target likelihood function of state estimation can be described as follows:

\begin{equation}
\arg \max P\left( {{x_k}\mid {y_{k + 1}}} \right) = P\left( {{d_k}\mid {y_{k + 1}}} \right)P\left( {{x_k}\mid {d_k},{y_{k + 1}}} \right)
\label{eq2-1}
\end{equation}

Here, ${d_k}$ is the latent variable that is unobserved. As shown in (\ref{eq1}), state and input both have an implicit relationship with the current observation, which means the traditional kalman filter cannot be performed well due to the interplay between state and input. 

Therefore, Expectation-Maximum (EM) algorithm, an iterative method, to find the maximum likelihood or maximum a posteriori (MAP) estimates of parameters is naturally an ideal solution to this problem. 

\textit{E step}: E step is that given the observation ${y_{k+1}}$ and last estimated state ${x_{k}}$ to determine the distribution of the parameter ${d_k}$, 

\begin{equation}
Q({d_k}) = {E_z}[P(z_k,{d_k}|{y_{k + 1}})]
\label{eq2-2}
\end{equation}

where $z_k$ is the latent value given the last state, input estimation, and observation. In the KF background, the innovation can be regarded as the $z_k$.

\textit{M step}: M step is to find the MAP estimation of input ${d_{k}}$ of $Q({d_k})$:
\begin{equation}
\mathop {\arg }\limits_d \max Q({d_k})
\label{eq2-3}
\end{equation}

It is deserved to claim that, if multiple timesteps delay is allowed, the EM process will run after the system accumulates batch observation data by a time window. If the online solution is required, EM can only loop one time since sequential data is not available. In the following chapters, we give a one-timestep online solution as an example to present the concurrent estimation by EM algorithm. To be specific, E step is using the innovation to find the optimal estimation of ${d_k}$:


\begin{equation}
\hat{x}_{k \mid k-1}=A_{k-1} \hat{x}_{k-1 \mid k-1}
\label{eq3}
\end{equation}
\begin{equation}
\hat{d}_{k-1}=M_{k}\left(y_{k}-C_{k} \hat{x}_{k \mid k-1}\right)
\label{eq4}
\end{equation}
where superscript of $\hat{x}_{k \mid k-1}$ means the estimated value. $M_k \in {\mathbb{R}^{m\times p}}$ is the input estimation gain. The specific format of the gain matrix $M_{k}$ depends on whether the parameter space is continuous or discrete space.

M step is then performed by Kalman filter to find the expectation of ${x_{k}}$ based on MVU principle.
\begin{equation}
\hat{x}_{k \mid k}^{\star}=\hat{x}_{k \mid k-1}+G_{k-1} \hat{d}_{k-1}
\label{eq5}
\end{equation}
\begin{equation}
\hat{x}_{k \mid k}=\hat{x}_{k \mid k}^{\star}+K_{k}\left(y_{k}-C_{k} \hat{x}_{k \mid k}^{\star}\right)
\label{eq6}
\end{equation}

where $K_k \in {\mathbb{R}^{n\times p}}$ are the state estimation gain matrix. 

The complete statement of recursive format of joint state and input estimation is shown in (\ref{eq3}-\ref{eq4}), (\ref{eq5}-\ref{eq6}). However, the contribution of this paper differs from previous research in that to derive the optimal input estimation on the condition of prior limitation but won't spoil the Gaussian distribution assumption of the system. Moreover, we not only consider the continuous space but also analyze the asymptotically unbiased estimation under discrete circumstances of input.



\section{One-timestep Unbiased and Minimum-variance of Input and State Estimation}
\label{section3}
This section gives a one-timestep EM algorithm as an example. The derivation of MVU estimation is presented without any limitation of input distribution. 

\subsection{Expectation (E) Step: Input $d_k$ Estimation}
\subsubsection{Unbiased Input Estimation}

From the system model in (\ref{eq1}) and (\ref{eq2}), the observation ${y_k}$ follows that 
\begin{equation}
    y_k = {C_k}({A_{k-1}}{x_{k - 1}} + {G_k}{d_{k - 1}} + {w_k}) + {v_k}
\label{eq7}
\end{equation}

Substituting (\ref{eq7}) into (\ref{eq4})
\begin{equation}
    {\hat d_{k - 1}}\! =\! {M_k}{C_k}{G_{k - 1}}{d_{k-1}}\! -\! {M_k}{C_k}{A_{k - 1}}({x_{k - 1}}\! -\! {\hat x_{k - 1}}) + {M_k}({{C_k}{w_k}+v_k})
\label{eq8}
\end{equation}

The subscript hat of ${\hat d_{k - 1}}$ means estimated value, while $d_{k-1}$ without subscript represents the groundtruth. Suppose at the last timestep ${\hat x_{k - 1}}$ is unbiased and $w_k$ is an mean-zero Gaussian noise. An unbiased estimation of input $d_{k-1}$ is obtained after finding the expectation of (\ref{eq8}) when satisfy

\begin{equation}
{M_k}{C_k}{G_{k - 1}} = {I_m}
\label{eq9}    
\end{equation}

with the rank limitation in \cite{gillijns2007unbiased} that 
\begin{equation}
    rank({C_k}{G_{k - 1}}) = rank({G_{k - 1}}) = m
\label{eq10}
\end{equation}

\subsubsection{Minimum-variance Input Estimation}
\ 

Defining the innovation ${\tilde y_k}$

\begin{equation}
\begin{aligned}
{\tilde y_k} &= {y_k} - {C_k}{\hat x_{k|k - 1}} \\
&= {C_k}{G_{k - 1}}{d_{k - 1}} + {e_k} 
\end{aligned}
\label{eq13}
\end{equation}

where ${e_k}$ is given by 
\begin{equation}
    {e_k} = {C_k}({A_{k - 1}}({x_{k - 1}} - {\hat x_{k - 1|k - 1}} )+ {w_{k - 1}}) + {v_k}
\label{eq14}
\end{equation}

From (\ref{eq4}), the input estimation as follows
\begin{equation}
\begin{aligned}
    \hat{d}_{k-1} &=M_{k}\left(y_{k}-C_{k} \hat{x}_{k \mid k-1}\right) \\
    &={M_{k}}{C_k}{G_{k - 1}}{d_{k - 1}} + {M_{k}}{e_k}
\end{aligned}
\label{eq15}
\end{equation}

Under the condition of (\ref{eq9}), (\ref{eq15}) can be rewrited as
\begin{equation}
    \hat{d}_{k-1} ={d_{k - 1}} + {M_{k}}{e_k}
\end{equation}

Therefore, the  covariance matrix of parameter $\hat{d}_{k-1}$ is
\begin{equation}
\begin{aligned}
Cov(\hat{d}_{k-1}) &= E({\hat d_{k - 1}}\hat d_{k - 1}^T) \\
    &= E[({M_{k}}{e_k})({M_{k}}{e_k})^T] \\
    &= {M_k}{S_k}{M_{k}^T}
\end{aligned}
\label{eq17}
\end{equation}

where the $S_k$is the covariance matrix of (\ref{eq14}).



Hitherto, the target that to find the MVU estimation of input has been transferred into finding a $M_k$ to a minimum variance of (\ref{eq17}) under the condition of (\ref{eq9}). 

The mathematical function of such equality-constrained optimization problem is expressed as follows
\begin{equation}
    \begin{array}{cc}
         min & Tr({M_k}{S_k}{M^T}) \\
         s.t. & {M_k}{C_k}{G_{k - 1}} = I_m
    \end{array}
\label{eq21}
\end{equation}

After conducting the Lagrange multiplier equation, the solution of $M_k$ as follows


\begin{equation}
{M_k} = {[{({C_k}{G_{k - 1}})^T}{S_k^{ - 1}}({C_k}{G_{k - 1}})]^{ - 1}}{({C_k}{G_{k - 1}})^T}S_k^{ - 1}
\label{eq27}
\end{equation}

Equation (\ref{eq27}) is gain matrix that guarantees the MVU estimation of the input.
The covariance of corresponding input estimation based on (\ref{eq17}) is given by
\begin{equation}
    Cov(d_{k})={[{({C_k}{G_{k - 1}})^T}{({C_k}{P_{k|k - 1}}C_k^T + {R_k})^{ - 1}}({C_k}{G_{k - 1}})]^{ - 1}}
\end{equation}


\subsection{Maximum (M) Step: State ${x_k}$ Estimation}
In this chapter, we will introduce potential limitation of input, which would be a form of constraint into the optimization problem in (\ref{eq21}). Afterward, the equality constraint question would be converted into an inequality constraint problem. Moreover, based on the continuous or discrete distribution of input, different solutions are proposed.
\subsubsection{Unbiased State Estimation }
\ 
From (3)-(6) and defining the estimation error of (\ref{eq5}) as $\tilde x_k^* = {x_k} - \hat x_{k|k}^*$

\begin{equation}
\tilde x_k^* = {A_{k - 1}}{\tilde x_{k - 1}} + {G_{k - 1}}{\tilde d_{k - 1}} + {w_{k - 1}}
\label{eq11}
\end{equation}

where $\tilde x$ and $\tilde d$ are estimation error as definition as $\tilde x_k^*$. Suppose ${\hat x_{k - 1}}$ is unbiased, while the unbiased initialization is discussed in \cite{kong2020filtering}, and $d_{k-1}$ is unbiased when satisfy the assumption of (\ref{eq9})
\begin{equation}
    \mathbb{E}(\tilde x_k^*) = 0
\end{equation}

Naturally, the expectation of $\hat{x}_{k \mid k}$ of (\ref{eq6}) is $x_k$. Thus, the unbiased state estimation is demonstrated.

\subsection{Minimum-variance State Estimation }

Combining (\ref{eq3} - \ref{eq6}) and (\ref{eq13}) together and then the final state estimation is



\begin{equation}
\begin{aligned}
    {\hat x_{k|k}} &= {\hat x_{k|k - 1}} + {K_k}{\tilde y_k} + ({I_n} - {K_k}{C_k}){G_{k - 1}}{M_k}{\tilde y_k}\\
    &= {\hat x_{k|k - 1}}  + {L_k}{\tilde y_k}
\end{aligned}
\label{eq30}
\end{equation}

\cite{gillijns2007unbiased} demonstrated that the best $K_k$ is a function of $M_k$ and varied by the rank of $y_k$ and $x_k$. 

\section{Prior Knowledge based Input Estimation}
\label{section4}

In real cases, prior input information is partially known. For instance, in maneuver tracking, the motion space of input is limited to a finite space; or in environment monitoring, classifications of earthquake are constrained by the discrete levels which play as a strong reference in estimation. Such prior knowledge serves as an extra constraint that would limit the range of inputs in optimization. 
This chapter mainly presents to the improvement of E step based on continuous space and discrete space with prior knowledge, M step is the same as Section III, which won't be repeated here.

\subsection{Continuous Space Constraint}

From (\ref{eq4}), suppose ${d_k}$ has a prior continuous limitation that ${d_k}={M_k} \tilde y_k \in [{N_1},{N_2}]$, the original optimization problem (\ref{eq21}) is becoming into
\begin{equation}
    \begin{array}{cc}
         min & f(M)=Tr({M_k}{S_k}{{M_k}^T}) \\
         s.t. & h(M) ={M_k}{C_k}{G_{k - 1}} - I_m =0 \\
              & {g_1}(M) = M_k {\tilde y_k } \ge   {N_1} \\
              &  {g_2}(M) = M_k {\tilde y_k } \le   {N_2}
    \end{array}
\label{eq31}
\end{equation}

where $\tilde y_k $ is the same as (\ref{eq13}). The inequality constraint problem of (\ref{eq31}) cannot be directly solved by K.K.T. conditions and penalty function because condition number of the Hessian matrix and concurrent satisfaction of two inequality constraints (\ref{eq31}) cannot be guaranteed. Thus, we utilize Augmented Lagrangian Kalman filter (AL-RKF) instead as the solution.

Firstly, conducting new target function $\varphi {}_1(M,\mu,\sigma )$ and $\varphi {}_2(M,\lambda,\gamma,\sigma )$ , where $\varphi {}_1(M,\mu )$ represents the equality constraint as follows
\begin{equation}
    \varphi {}_1(M,\mu,\sigma ) = Tr(f(M) - \mu h(M) + \frac{\sigma }{2}{h^2}(M)) 
    \label{eq32}
\end{equation}

As for the inequality constraint of ${g_i}(M)$ in (\ref{eq31}), introducing parameter $\gamma$ to transfer the inequality constraint into equality constraint.

\begin{equation}
\begin{aligned}
        &\varphi {}_2(M,{\lambda _i},{\gamma _i},{\sigma}) \\&= Tr(f(M)\! -\! \mathop \Sigma \limits_{i = 1}^2 [{\lambda _i}({g_i}(M)\! -\! \gamma _i^2]\! +\! \frac{\sigma }{2}\mathop \Sigma \limits_{i = 1}^2 {({g_i}(M) - \gamma _i^2)^2}) \\
    & = Tr(f(M) + \mathop \Sigma \limits_{i = 1}^2 \left\{ {\frac{\sigma }{2}{{[\gamma _i^2 - \frac{1}{\sigma }(\sigma {g_i}(M) - {\lambda _i})]}^2} - \frac{{\lambda _i^2}}{{2\sigma }}} \right\})
\end{aligned}
\label{eq33}
\end{equation}

To minimize function (\ref{eq32}), ${\gamma_i}^2$ should follow the equation
\begin{equation}
    \gamma _i^2 = \frac{1}{\sigma }{[\max \{ 0,\sigma {g_i}(M) - {\lambda _i}\} ]^2}
\label{eq34}
\end{equation}



Then complete function of (\ref{eq31}) is
\begin{equation}
    \begin{split}
        \varphi (M,\mu ,{\lambda _i},\sigma ) = &f(M)- \mu h(M) + \frac{\sigma }{2}{h^2}(M) \\
        &+ \frac{1}{{2\sigma }}\mathop \sum \limits_{i = 1}^2 \{ {[\max \{ 0,{\lambda _i} - \sigma {g_i}(M)\} ]^2} - \lambda _i^2\} 
    \end{split}
\label{eq36}
\end{equation}

where parameters $\mu$ and ${\lambda}$ are multiplier which objectively have optimal values ${\mu ^ * }$ and ${\lambda ^ * }$; while parameter ${\sigma}$ is a penalty factor by artificial setting.
It is necessary to claim that the inequality optimization won't change the Gaussian property of (\ref{eq15}), since inequality functions only have an influence on the input gain $M_k$ which won't affect the Gaussian property of ${\hat d_{k - 1}}$. 

The multiplier improvement strategy as follows
\begin{equation}
\begin{aligned}
    \lambda _i^{(k + 1)} &= \max (0,\lambda _i^{(k)} - \sigma {g_i}(M)),i = 1,2,...,m \\
    {\mu ^{(k + 1)}} &= {\mu ^{(k)}} - \sigma h({M^k})
\end{aligned}
\label{eq37}
\end{equation}


Optimization process is as Algorithm \ref{algorithm1}.

\begin{algorithm}[htb]  
\caption{AL-RKF: Continuous Space Input Estimation}\label{algorithm1}
\SetAlgoLined
\SetKwData{Left}{left}\SetKwData{This}{this}\SetKwData{Up}{up}
\SetKwFunction{Union}{Union}\SetKwFunction{FindCompress}{FindCompress}
\SetKwInOut{Input}{input}\SetKwInOut{Output}{output}

\Input{ Given system model $A_k$, $C_{k+1}$, $G_k$, $Q_k$, $R_k$, last state $x_k$, $P_{k}$, observation ${y_k}$, and prior scope limitation of input $[{N_1, N_n}]$}
\Output{The optimal estimation of input $d_k$ and state $x_{k+1}$}

Let $M_0\leftarrow (\ref{eq27})$ and conducting target optimization function $f(M)$, $h(M)$, $g(M)$ as Eq.(\ref{eq31}) \\ 
    \eIf{$M_0 {\tilde y_k} \ge  {N_1}$ \&  $M_0 {\tilde y_k} \le  {N_n}$}{
      $d_k = M_0 {\tilde y_k k}$\; 
      }{
      Set initial $\sigma$ to a large number, $\lambda=0$, $\mu=0$ and step $k=0$\; \\
      \While{$(M_{k+1}-M_k \ge \delta)$}{

        ${M^{(k + 1)}} = \arg \mathop {\min }\limits_M \varphi (M,\mu ,{\lambda _i},\sigma )$; see Eq.(\ref{eq36}) \; 
        $ \lambda _i^{(k + 1)} = \max (0,\lambda _i^{(k)} - \sigma {g_i}(M))$, i = 1,2 \; \\
        ${\mu ^{(k + 1)}} = {\mu ^{(k)}} - \sigma h({M^k})$ \;\\
        $k= k+ 1$ \; 
      }
}
\end{algorithm}

\subsection{Discrete Space Constraint}
In this section, suppose ${d_k}={M_k} \tilde y_k $ is in a set of discrete modes ${ \{{N_1},{N_2},...,{N_n}\}}$, in which the discrete attributes could provide a reference to realize a higher accurate estimation, an adaptive Multi-modes Kalman Filter (AMM-KF) algorithm is then proposed. Due to discrete knowledge, estimation of inputs is not directly obtained by MAP estimation. Actual decisions are made by probability density function (PDF) of potential modes. Soft-decision is introduced to reduce the bias of hard-decision. Asymptotically unbiased is also demonstrated.

Mathematically, the optimization problem (\ref{eq21}) is
\begin{equation}
    \begin{array}{cc}
         min & f(M)={M_k}{S_k}{{M_k}^T} \\
         s.t. & {M_k}{C_k}{G_{k - 1}} - I_m =0 \\
              & M_k {\tilde y_k} \in  \{N_1, N_2, ..., N_n\}
    \end{array}
\label{eq38}
\end{equation}


In practice, If input space is limited to discrete modes, a more efficient solution than continuous space  to (\ref{eq38}) is available. 

\begin{algorithm}[htb]  
\caption{AMM-KF:Discrete Space Input Estimation}\label{algorithm1}
\SetAlgoLined
\SetKwData{Left}{left}\SetKwData{This}{this}\SetKwData{Up}{up}
\SetKwFunction{Union}{Union}\SetKwFunction{FindCompress}{FindCompress}
\SetKwInOut{Input}{input}\SetKwInOut{Output}{output}

\Input{ Given system model $A_k$, $C_{k+1}$, $G_k$, $Q_k$, $R_k$, last state $x_k$, $P_{k}$, observation ${y_k}$, and prior scope limitation of input ${\{{N_1},{N_2},...,{N_n}\}}$}
\Output{The right decision-making of input $d_k$}

Initialize the weights of $\mu$ with equal probability of each modes.
\\ 
    \eIf{max $\mu_{k}^{(i)}$ is $N_i$ $\ge 90\%$ {\rm time} }{
      max $\mu_{k}^{(i)}$ = $N_i$ ; 
      }{    
       \rm Find the likelihood functions of different modes \\ 
        $L_k^i = N({\tilde y_{k + 1}}\;{\mu _{{d_i}}},{\Sigma _{{d_i}}})$  \\
        \rm Update the weights $\mu$ of estimator  \\
        $ \mu_{k}^{(i)}=\frac{\mu_{k-1}^{(i)} L_{k}^{(i)}}{\sum_{j=1}^{n} \mu_{k-1}^{(j)} L_{k}^{(j)}}$ \, i = 1,2, ...  \\
        \rm Make the decision of $d_k$ at one timestep \\
        ${d_k} = \mathop {\max }\limits_{{d_i}} (L_k^i)$ \\
        \rm Update the state ${\hat x_{k + 1|k + 1}}$and covariance ${P_{k + 1\mid k + 1}}$ in Eq.(\ref{eq42}) and Eq.(\ref{eq43}); 
     }
\end{algorithm}
 


Simplicity, we take two modes ${N_1, N_2}$ as an example. Recalling the standard Kalman Filter without information of input $d_k$. The final updated equation is
\begin{equation}
    {x_{k + 1|k + 1}} = {A_k}{x_{k|k}} + {K_k}({y_{k + 1}} - {C_k}{A_k}{x_{k|k}})
\label{eq39}
\end{equation}

Suppose input can be precisely estimated, the ensemble filter combing two filters is
\begin{equation}
\begin{split}
    {\hat x_{k + 1|k + 1}} =&{A_k}{\hat x_{k|k}} + {K_k}({y_{k + 1}} - {C_{k + 1}}{A_k}{\hat x_{k|k}}) \\
    &+ (I - K{C_{k + 1}}){G_k}\mathop \Sigma \limits_{i = 1}^2 ({\mu _i}d_k^i)
\end{split}
\label{eq42}
\end{equation}

where $\mu$ is the weight of different modes. Covariance matrix is then updated by 

\begin{equation}
{P_{k + 1\mid k + 1}} = \sum\limits_{i = 1}^N {\left[ {P_{k + 1\mid k + 1}^{(i)} + e_{k + 1}^ie{{_{k + 1}^i}^T}} \right]} \mu _{k + 1}^i
\label{eq43}
\end{equation}

where $P_{k\mid k}^{i}$ is the covariance of a single sub-filter. The error item $e_{k + 1}$ is the  ${\left( {{{\hat x}_{k + 1\mid k + 1}} - x_{k + 1\mid k + 1}^{(i)}} \right)}$.

Therefore, the bullet point is to find a parameter $\mu$ in the error term $(I - K{C_{k + 1}}){G_k}\mathop \Sigma \limits_{i = 1}^2 ({\mu _i}d_k^i)$.


Define the likelihood function as

\begin{equation}
    L_k^i = N({\tilde y_{k + 1}};{\mu _{{d_i}}},{\Sigma _{{d_i}}})
\label{eq44}
\end{equation}

The input and weight are determined by following  

\begin{equation}
    {d_k} = \mathop {\max }\limits_{{d_i}} (L_k^i) 
\label{eq45}
\end{equation}
\begin{equation}
    \mu_{k}^{(i)}=\frac{\mu_{k-1}^{(i)} L_{k}^{(i)}}{\sum_{j=1}^{n} \mu_{k-1}^{(j)} L_{k}^{(j)}}
\label{eq46}
\end{equation}

where the parameter $\mu$ is the dynamic weight of input compensation term in (\ref{eq42}) which can dynamically updated along with observation. Moreover, the weight $\mu$ can strongly cope with outliers in case large variance happens. 

However, the convergence efficiency of $\mu$ highly depends on the property of system models. The key is that the separation between different modes should be identifiable enough. If the value of inputs isn't comparable to the process noise, then discrimination would be failed. 

From the knowledge of hypothesis testing, the positive testing probability is 
\begin{equation}
         P_D = P({H_0}|{H_0})=\int_{\lambda}^{\infty}PDF_{H_1}(U)=\int_{\lambda}^{\infty}{N}(\mu_{u^1},\Sigma_{u^1}) 
 \label{eq47}
\end{equation}

where $\lambda$ is the intersection boundary when ${N}(\mu_{u^2},\Sigma_{u^2}) = {N}(\mu_{u^1},\Sigma_{u^1}) $,  $P({H_0}|{H_0})$ means the right decision probability of when $H_0$ is successfully detected. $P_D$ sets a bar to system that how confidence of the decision at one timestep. The proportion of $P_D$ of multiple modes is the parameter $\mu$. With sequential observation, $\mu$ will converge to the right mode gradually. Therefore, it is asymptotically unbiased.

\section{Simulation and Experiment Result}
\label{section5}
\subsection{Experiment Setup}

\begin{figure}[htb]        
 \center{\includegraphics[width=8.6cm]  {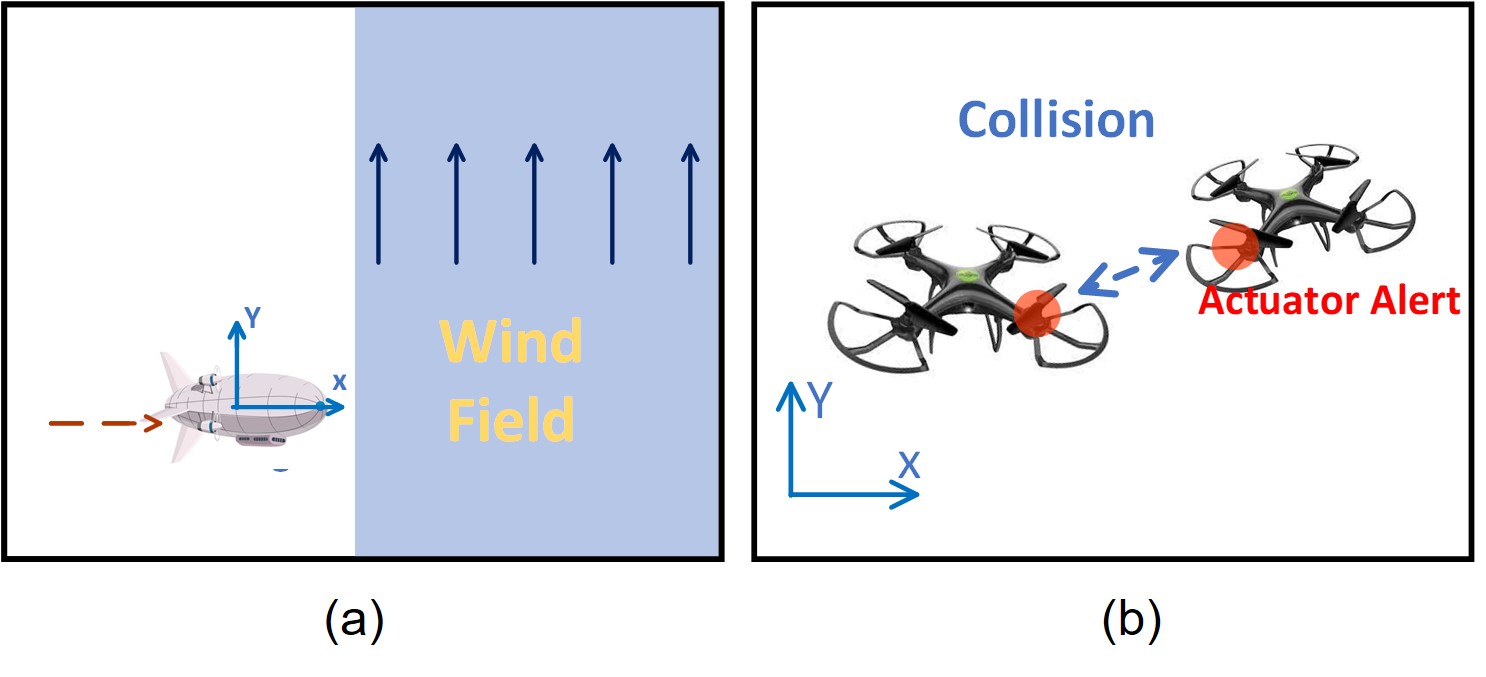}}     
 \caption{\label{demo}(a) A airship is flying forward into a wind field. (b) In swarm system, multi-agents may have collision with others. Actuator might failed after one agent collides with others.}
\setlength{\belowcaptionskip}{-1cm}  
\vspace{-0.2cm} 
\end{figure}

To intuitively and numerically demonstrate our algorithm, two examples as Fig.\ref{demo} are created to illustrate potential applications of the algorithm, which cover the continuous case and discrete case, respectively. 

One case is to simultaneously estimate the position of airship and wind speed without an additional anemograph sensor. The prior knowledge is that the wind speed is limited between $0 m/s$ and $10.3 m/s$ with the direction to $y$ axis. During this process, we suppose the mass of airship $m$, air mass density $\rho$ and superficial area $S$ of the airship acted by wind force are all known. Trajectory observation coming from satellites but with noise. The motion model used in KF is Newton's kinematics equations. The goal is that using observations from satellite to estimate the accurate position of airship as well as the wind speed. 

Another case is in swarm systems where multiple agents are independently running. One drone is hovering but then collides with adjacent one. Collision will make one actuator failed. The assumption is that the collision and actuator failure only affects the horizontal location of drones. Observations are coming self-localization with noise, which leads to a deficiency gain values $K$ in Eq.(\ref{eq39}). The goal is to estimate the probability of actuator failure by collision occurrence. The prior knowledge is that failure influence is discrete, namely collision occurrence is binary. By this algorithm, users might consider recalling this agent back to repair shop based on the status of collision evaluation in real applications.

\subsection{Result and Discussion}
\begin{figure}[htb]        
 \center{\includegraphics[width=8.6cm]  {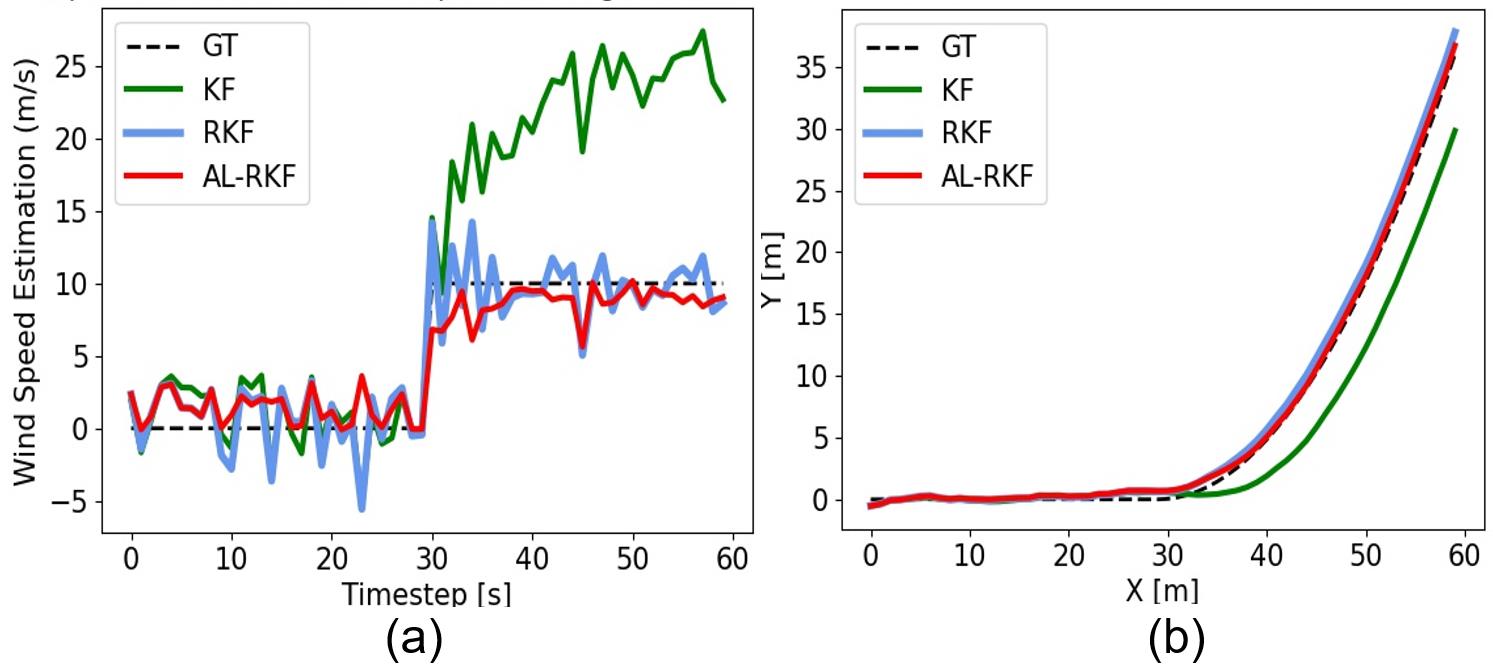}}   
 \caption{\label{rkf} The airship moved into wind field at 30s, which corresponding location is (30m,0m). (a) External wind force $F$ is directly estimated by filter, then wind speed is calculated by equation $F = \rho S{v^2}$. (b) The x and y axes represent the coordinates in the geometrical space. Four trajectories represents comparison between three filters and groundtruth.}
\setlength{\belowcaptionskip}{-1cm}  
\vspace{-0.2cm} 
\end{figure}

\begin{table}[htbp]
\setlength{\abovecaptionskip}{0.cm}
\setlength{\belowcaptionskip}{-0.1cm}
\renewcommand{\arraystretch}{1.3}
\caption{ Statistical Result of Estimation Error }
\label{table1}
\begin{center}
 \resizebox{8.6cm}{12.5mm}{
\begin{tabular}{|c|c|c|c|c|c|c|c|}
\hline
\multicolumn{2}{|c|}{\textbf{Q}}&\multicolumn{3}{|c|}{\textbf{[0.05,0.05]}}
&\multicolumn{2}{|c|}{\textbf{[0.05,0.5]}}
&{\textbf{[0.5,0.5]}}\\
\hline
\multicolumn{2}{|c|}{\textbf{R}}
&{ \textbf{{[0.05,0.05]}}}
&{ \textbf{{[0.05,0.5]}}} 
&{ \textbf{{[0.5,0.5]}}}
&{ \textbf{{[0.05,0.5]}}}
&{ \textbf{{[0.5,0.5]}}}
&{ \textbf{{[0.5,0.5]}}}\\
\hline
&KF&2.84&2.5&6.4&2.02&6&14.29\\
\cline{2-8}
\multirow{2}{*}[4pt]{\shortstack{mean\\(m)}}&RKF &\textbf{0.06}&\textbf{0.4}&0.5&0.27&\textbf{0.37}&\textbf{0.13} \\
\cline{2-8}
&AL-RKF&0.08&0.45&\textbf{0.25}&\textbf{0.2}&0.41&0.34\\
\hline
&KF&7.7&640&80&17.5&148&111.4\\
\cline{2-8}
\multirow{2}{*}[4pt]{\shortstack{variance\\($m^2$)}}&RKF &0.8&5.0&6.3&14.2&148&69.8 \\
\cline{2-8}
&AL-RKF&\textbf{0.3}&\textbf{3.2}&\textbf{4.1}&\textbf{9.0}&\textbf{25}&\textbf{48}\\
\hline
\end{tabular}}
\label{tab1} 
\end{center}
\vspace{-0.4cm} 
\end{table}


Fig.\ref{rkf} shows that both RKF and AL-RKF obviously perform better than KF when external input acts on dynamic systems. The primary advantage of AL-RKF compared to RKF is that we use prior knowledge as limitation to reduce the variance of input estimation. To thoroughly explore the system noise influence on input estimator, multiple experiments with different couples of $Q$ and $R$ are carried out. Table \ref{table1} shows the mean and variance of estimation error. Statistical results indicate that, on average, AL-RKF has 92\% improvement of mean error than kF only with relatively minor aggravation than RKF, but AL-RKF owns 81\% improvement of the variance than KF and 47\% improvement than RKF.

The reason for unnotable trajectories comparison can be explained by (\ref{eq39}) and (\ref{eq42}). If the values of input bias and state are not at the same orders of magnitude, state estimation comparison then wouldn't show a significantly difference.

\begin{figure}[htb]        
 \center{\includegraphics[width=8.6cm]  {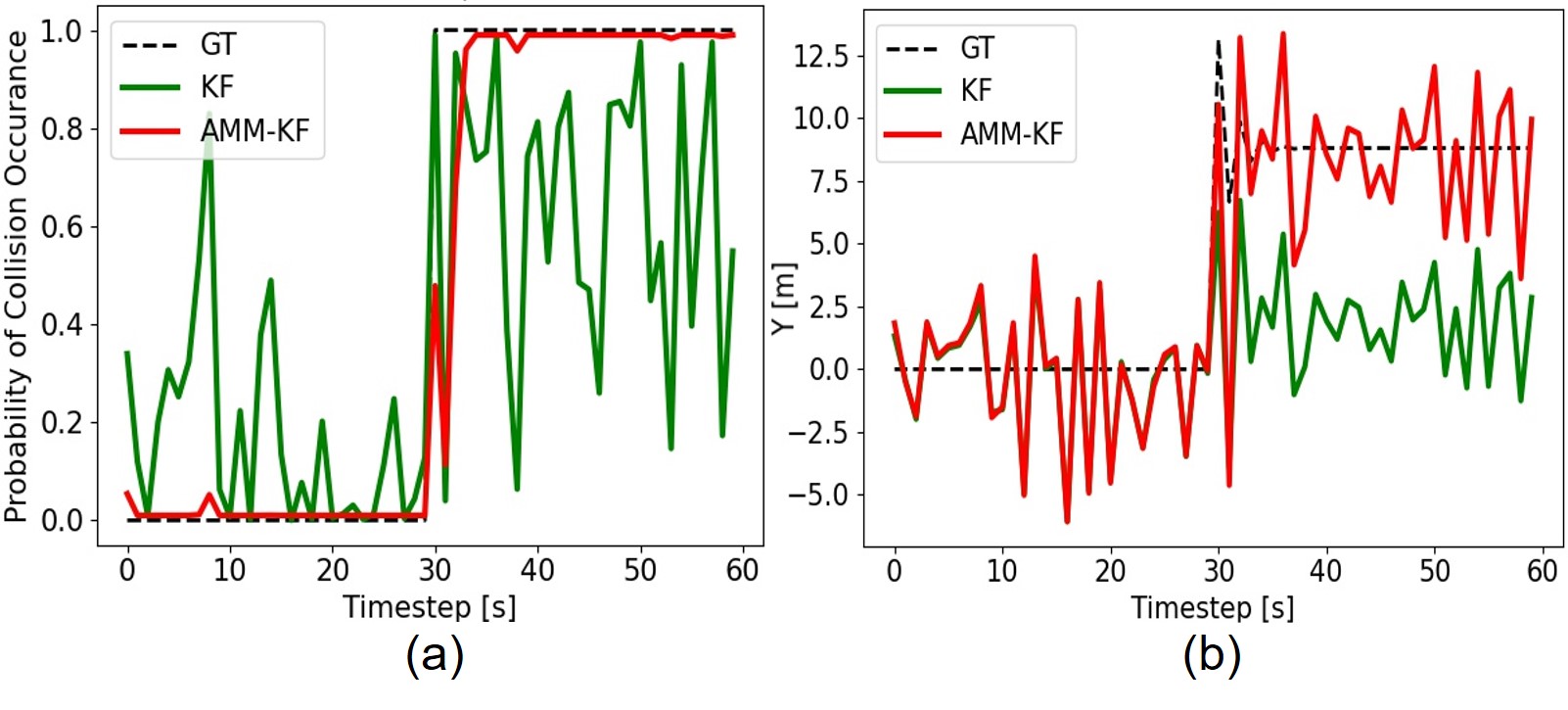}}   
 \caption{\label{mm1} The sensor failed caused by collision at 30s. (a) The y axis represents the confidence about the sensor failure. (b) The y axis represents displacement at y direction of space. }
\setlength{\belowcaptionskip}{-5cm}  
\vspace{-0.2cm} 
\end{figure}

Fig.\ref{mm1} shows the probability of sensor failure estimation. The probability of AMM-KF is shown by dynamical parameter $\mu$ in Eq.\ref{eq46} while KF is the direct hypothesis testing estimation by probability $P_D$ in Eq.\ref{eq47}. The input decision is made when probability is larger than 0.5. This figure shows that through sequential observation, the confidence of inputs estimation stably maintained to the right decision while standard KF has large variance. As for the position estimation, since we precisely estimated the failure, compensation from observation is successfully fit in the bias of standard KF. Therefore AMM-KF is more significant than KF when the actuator failed.

\begin{figure}[htb]        
 \center{\includegraphics[width=8.6cm]  {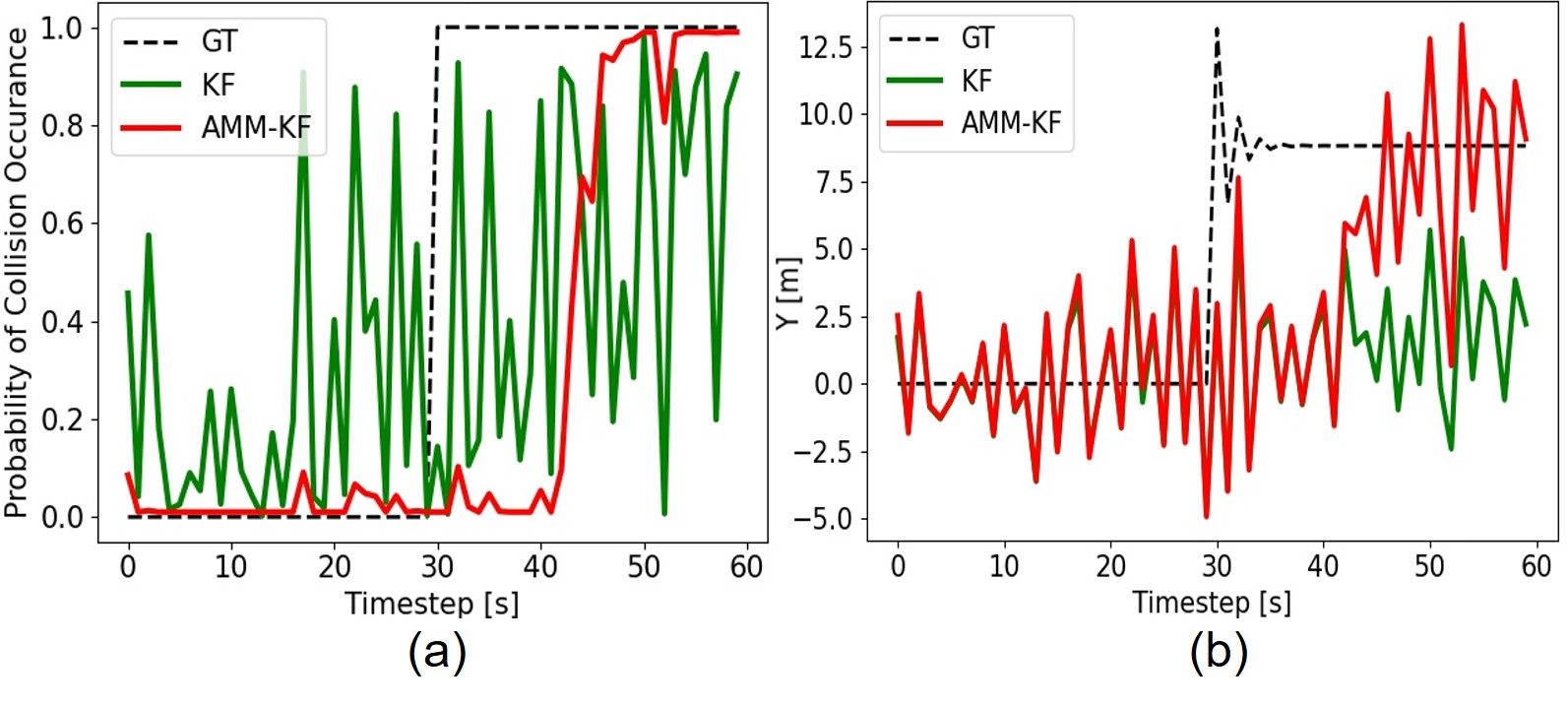}}   
 \caption{\label{mm1-2} Input estimator not always converges to right decision-making rapidly given large noise of system.}
\setlength{\belowcaptionskip}{-1cm}  
\vspace{-0.2cm} 
\end{figure}
Although input estimator is asymptotically unbiased, the convergence of decision-making are not always efficient enough. As shown in Fig.\ref{mm1-2}, system with large noise needs more time to converge. In practice, convergence efficiency of estimator directly affect practical usage, otherwise unstable and long-time convergence will consume too many resources.

From the comparison between Fig.\ref{mm1-2} and Fig.\ref{mm1}, we found that the probability of decision is closely related to the one-timestep hypothesis testing probability $P_D$. The mean of the Gaussian distribution in Eq.\ref{eq47} is the innovation from  Eq.(\ref{eq13}) and the corresponding variance is Eq.(\ref{eq17}). During the estimation process, innovation and variance dynamically change which is reflected by the fluctuated probability curve of KF, and thus affect the actual decision-making.
\begin{figure}[h]        
 \center{\includegraphics[width=0.52\textwidth] {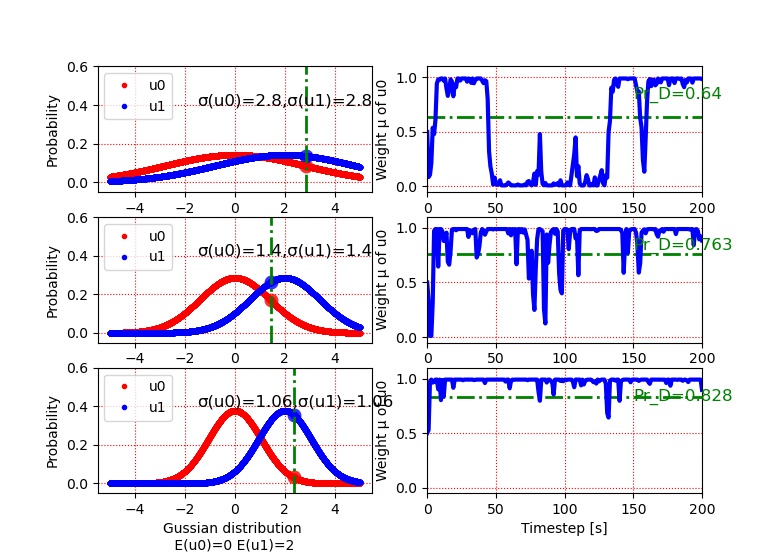}}
 \caption{ \label{mm3} Dynamic updating process of parameter $\mu$ given two different Gaussian distributions. To avoid decision-making ambiguity, separation between different distributions should be identifiable enough.}
\setlength{\belowcaptionskip}{-1cm}  
\vspace{-0.2cm} 
\end{figure}

In order to examine the influence of innovation and variance on the input estimator, a simulation is carried out where two Gaussian distributions with different mean values are given. Fig.\ref{mm3} shows the dynamic process of the estimator. Mean of Gaussian distribution is similar to innovation in Eq.(\ref{eq13}), and $\sigma$ plays as the variance in Eq.(\ref{eq17}). Green line is the $P_D$ calculated by Eq.(\ref{eq47}). Results show that the bottom distribution where both $\sigma=1.06$ gives the right decisions more than 90\% time while the top distribution with $\sigma=2.8$ makes the wrong decision more than 50\% time. Therefore, when the innovation and variance is like the bottom distribution in Fig.\ref{mm3} , the estimator can always make the right decision. However, if innovation and variance fit the top distribution, our estimator might fail with a higher possibility.

\section{Conclusion}
\label{section7}

This paper explored a novel recursive filter based on EM algorithm to simultaneously estimate the state and input of dynamic systems. Distinct from previous research, this paper considered a prior knowledge as a constraint to facilitate input estimation. In practice, inputs could be continuous, changing smoothly in continuous space, or discrete, only changing through discontinuous ``jumps", which are both discussed in our paper.

Experiments in this paper have given intuitive examples of applications of our algorithm. More quantitative analysis of convergence will be further studied in the future work. The value of our algorithm is to better evaluate the unknown disturbance on a dynamical system, which can extend previous state estimator capability without adding additional sensors. Unknown inputs could be an environmental disaster event, internal sensor failure, or an external force by collision, in addition to others, depending on the specific dynamic system.

\section*{Acknowledgement}
This work has been funded by DARPA grant HR00112090027.
\addtolength{\textheight}{-12cm}   

\bibliographystyle{ieeetr}
\bibliography{ref} 
\end{document}